\documentclass[doublespaced, 14pt]{ut-thesis}

\usepackage{hyperref}
\usepackage{graphicx}

\usepackage{tipa}
\usepackage{booktabs}
\usepackage{adjustbox}
\usepackage{fancyhdr}
\usepackage{tabularx,ragged2e}
\usepackage{longtable}
\usepackage{enumerate}
\usepackage{amsmath}
\usepackage{color,soul}
\usepackage[table]{xcolor}
\usepackage{pagecolor}

\newcolumntype{C}{>{\Centering\arraybackslash}X} 
\newcolumntype{L}[1]{>{\hsize=#1\hsize\raggedright\arraybackslash}X}%
\newcolumntype{G}[1]{>{\hsize=#1\hsize\centering\arraybackslash}X}%

\hypersetup{colorlinks=true, citecolor=green, linkcolor=red}

\degree{Master of Science in Computer Science}
\department{Computer Science}
\gradyear{2018}
\author{Nancy Iskander}
\title{Generating 3D models from sketches of human faces using 
a combined approach of Convolutional Neural Networks, Procedural Modeling,
and Contour Mapping}

\begin{document}

\begin{preliminary}

\maketitle


\begin{abstract}
Generating 3D models from face sketches is an active topic of research
in Computer Graphics due to its potential to tremendously facilitate 
the modeling of faces for both professional 3D arists and novices.

Motivated by the observation that facial expressions are responsible for 
significantly altering and shaping the contours in our faces, we combine 
both expression detection and 3D model generation in our approach. 
The result is a novel approach to generating 3D models from sketches which relies 
on three components: Convolutional Neural Networks, a parametric 
3D face model (Valley Girl), and Active 
Snake Contours. For the first time in the literature, 
CNNs are trained (using our own generated dataset) to detect the expression in the given 
sketch through detecting the active FACS Action Units. The expression 
is then duplicated on Valley Girl to obtain a 3D model with a similar 
expression. Active Snake Contours are then used to find the transforms 
needed to close the gaps between that model and the given sketch.

\end{abstract}





\begin{acknowledgements}
Thank you to my supervisor Karan Singh for going above and beyond in order to help me
grow as a researcher and as a person, for believing in me, and for general awesomeness. \\

Thank you to Chris Landreth for sharing his parametric face model (Valley Girl)
and for imparting his knowledge of FACS, Maya, and facial animation on me.  \\

Thank you to Sanja Fidler for detangling Deep Learning and making it seem accessible. \\

Thank you to Lynda Barnes for exceptional response times whenever I needed information
on the program.\\

Thank you to my parents Eman Soliman and Magdy Hanna, my brother Samir Iskander,
and Pierre-Luc Loyer for their support and motivation. \\

Finally, thank you to my employer Behaviour Digital for being accomodating of 
my schedule when I needed to work on this thesis.
\end{acknowledgements}

\tableofcontents

\listoftables

\listoffigures


\end{preliminary}


\chapter{Introduction}

Generating 3D models from line sketches is an interesting problem that has 
received a lot of attention in the literature, beginning with approaches that 
do contour matching between an existing model and the sketch~\cite{sketch2010}, 
to those  relying on deep learning to directly generate the model 
from the sketch~\cite{deepSketch}. This work develops a combined 
approach where both deep learning and contour matching between an existing
model and the sketch are used. 

\begin{figure}[ht]
\centering
\includegraphics[width=6in]{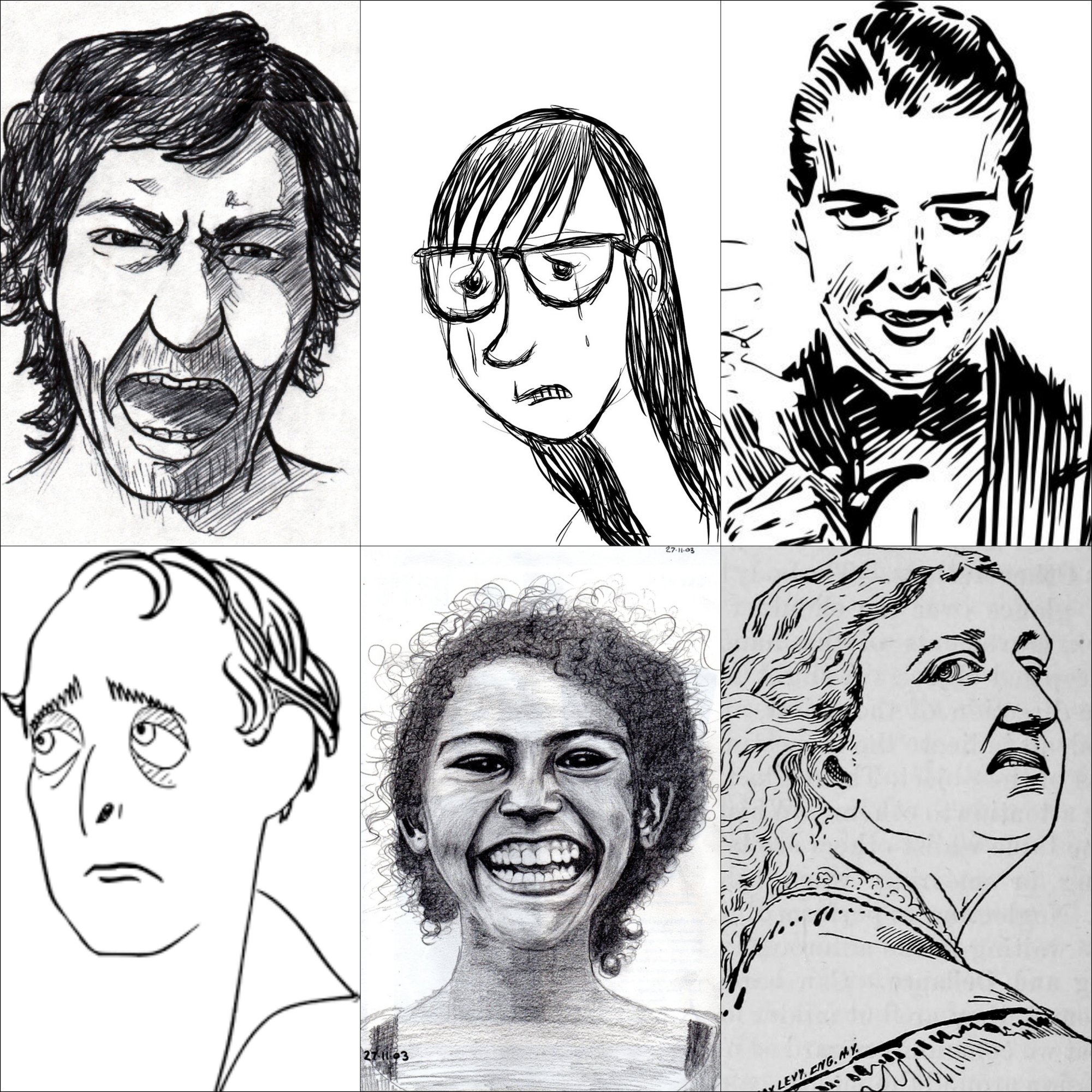}
\caption{Sketches of expressive faces found on the internet, where
a wide range of active AUs can be observed (e.g. ``smile'', ``brow furrow'',
``sneer'', etc).}
\label{fig:expressive}
\end{figure}

When drawn as a sketch, a face contorted into an expression will 
contain a greater number of contours than a blank face. Oftentimes, 
artists use a face sketch to portray a particular emotion, as seen
in figure~\ref{fig:expressive}. This is 
why we aim to detect and reproduce the expression from the given sketch 
on our model before attempting to do contour matching between them.
In order to detect the expression, we generated our own dataset of 
3D models and corresponding sketches, which is used to train 
Convolutional Neural Networks to detect facial actions in the given
sketch. 

\textit{Applications.} There is a wide range of applications for generating 3D 
models from sketches using our method in professional and non-professional settings: 
\begin{itemize}
\item 3D artists can quickly prototype their character designs and obtain a starting point for their model.
\item Novices and Indie developers can generate character heads with 
    little to no knowledge of 3D modelling software.
\item Witnesses and sketch-artists can obtain a higher level of accuracy by being able to quickly visualize 
      their sketches as renders and iterate on them.
\item Children can effortlessly see their simple sketches come to life.
\item An expression from a sketch can be replicated on any model.
\end{itemize}

Before delving into the details of our method, there are three components
that are important to understand: FACS (described in~\ref{sec:facs}), 
Valley Girl (described in~\ref{sec:vg}), 
and Snakes (described in~\ref{sec:snakesOverview}).

\section{Overview of FACS}
\label{sec:facs}
\begin{figure}[ht]
\centering
\includegraphics[width=3in]{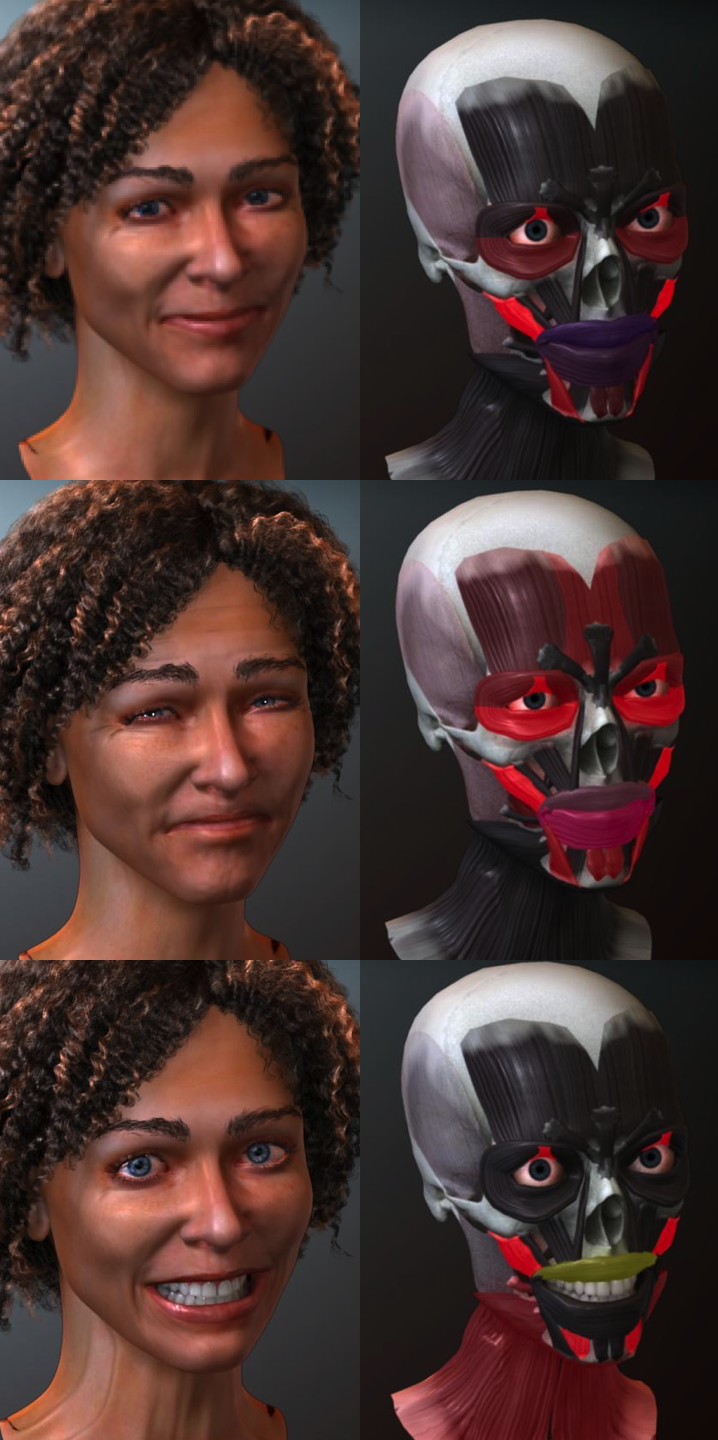}
\caption{Three facial expressions that can be construed as ``happy'' that are in fact very different.
The first column shows the expression on Valley Girl.
The second column shows which muscles are active in each expression.}
\label{fig:variation}
\end{figure}

The Facial Action Coding System (FACS) was developed by Ekman et al.~\cite{ekman1977facial}
in 1977. Describing the emotion portrayed by a facial expression (e.g. ``happiness'',
``sadness'', ``anger'', etc) is not precise enough and does not tell you what the facial 
expression actually looks like (see figure~\ref{fig:variation}). 
On the other hand, detecting the exact muscle groups which
are active in a facial expression is nearly impossible for both humans and computers 
(if the input is an image), because the relation between muscle actions 
and observable changes in facial appearance is many-to-many. 
For example, the \emph{frontalis} muscle can be used to raise one brow, to wrinkle the entire forehead, 
and to wrinkle just the center of the forehead (one muscle mapping to many facial changes). 
Conversely, the \emph{risorius} and \emph{platysma} muscles are always fired together and produce a 
grimace (many muscles mapping to one change).

FACS encodes only observable components of facial movement as AUs, and each facial 
expression can be broken down into its constituent AUs, which is what we aim
to vary in the generated dataset (section~\ref{sec:dataset}) and later detect (section~\ref{sec:training}).

\section{Overview of Valley Girl}
\label{sec:vg}
Valley Girl is a physiologically accurate 3D facial model developed 
in Maya by Chris Landreth~\cite{landreth}. All 
facial muscle actions are modelled using blendshapes and exposed 
as parameters modelled after FACS AUs. Table~\ref{mapping}
shows the mapping between a subset of Valley Girl facial actions,
muscles used, and FACS AUs.

The FACS AU parameters are divided into 12 upper-face controls, 
11 mid-face controls, and 24 lower-face controls. 
In addition, there are three controls for rotating the head 
in 3D space, as well as three for rotating the neck.
There are parameters for choosing the ethnicity and 
the gender which can be used to vary the facial features.
I have also augmeneted the original Valley Girl to add parameters
for varying facial features more explicitly (e.g. eye shape, 
jawline shape, etc).

\begin{figure}[ht]
\centering
\includegraphics[width=6in]{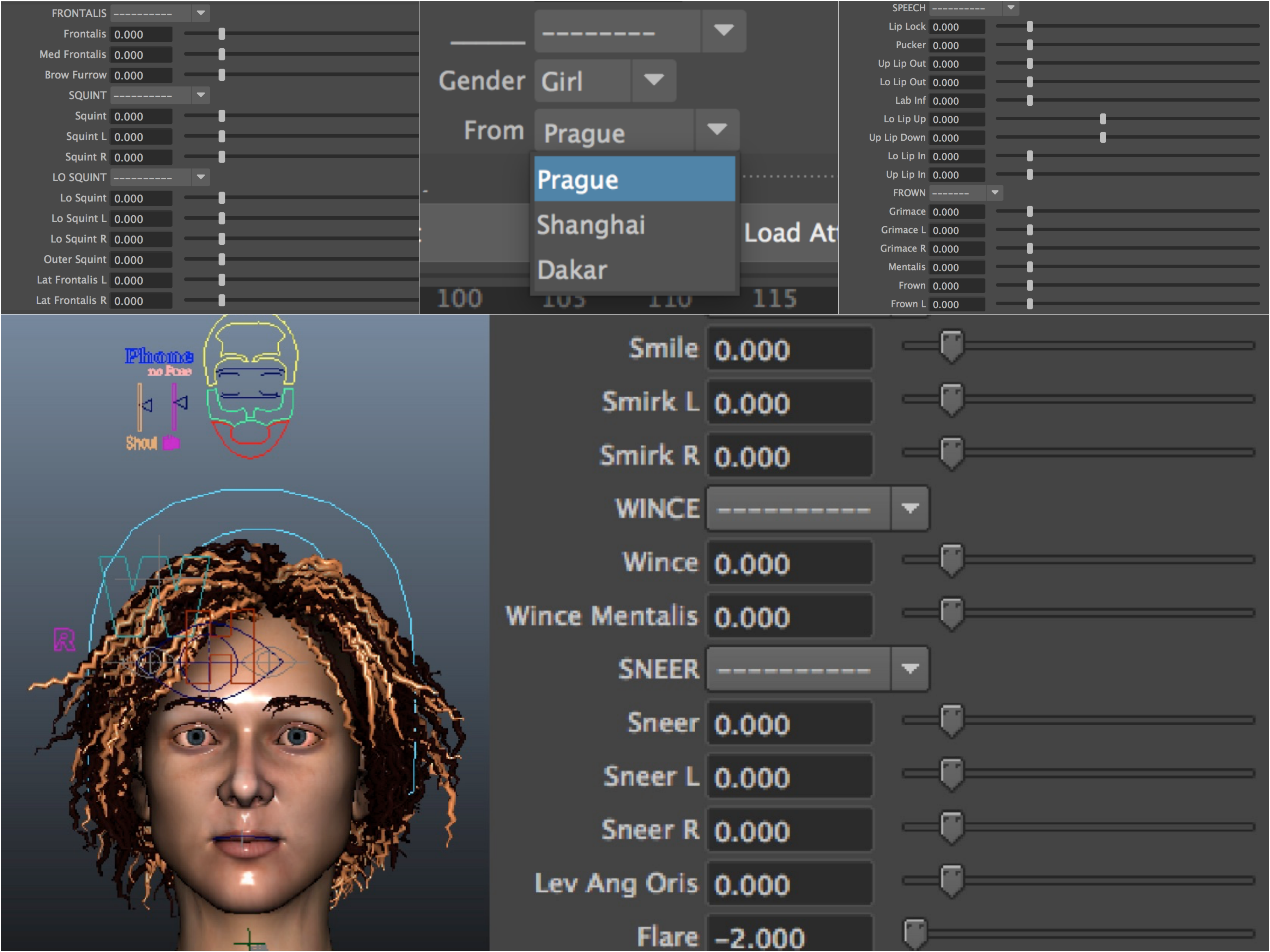}
\caption{Valley Girl's UI in Autodesk MAYA.}
\label{fig:ui}
\end{figure}

Figure~\ref{fig:ui} provides an idea of what Valley Girl's UI looks
like in Autodesk MAYA. These parameters can be used for posing
and facial animation. (The same parameters can be controlled 
programmatically.)

\begin{table*}[t]
\centering
\caption{Mapping between a subset of face actions used in Valley Girl UI, muscles used and FACS AUs.}
\label{mapping}
\begin{tabularx}{\textwidth}{CCCCC}
\toprule
\textbf{Action}    & \textbf{Signature}        & \textbf{Muscles Involved}                      & \textbf{FACS AUs} & \textbf{Valley Girl} \\ 
\midrule
Raising eyebrows   & Worry lines               & \textit{Frontalis}                             & 1, 2              &   \adjustimage{width=1.8cm,height=1cm,valign=m}{images/t1}          \\         
Lowering brow      & Furrowed eyebrows         & \textit{Currogator}                            & 4                 &   \adjustimage{width=1.8cm,height=1cm,valign=m}{images/t2}                 \\
Blinking           & Open/Closed eyes          & \textit{Levator palpebrae superioris}          & 5                 &    \adjustimage{width=1.8cm,height=1cm,valign=m}{images/t3}                  \\
Squinting          & Furrowed eyes             & \textit{Orbicularis Oculi}                     & 6, 44             &    \adjustimage{width=1.8cm,height=1cm,valign=m}{images/t4}                 \\
Wincing            & Squared lips, nasal folds & \textit{Levator Labii Superioris Alaeque Nasi} & 9                 &    \adjustimage{width=1.8cm,height=1cm,valign=m}{images/t5}       \\
Sneering           & Squared lips              & \textit{Levator Labii Superioris}              & 10                &          \adjustimage{width=1.8cm,height=1cm,valign=m}{images/t6}           \\
Smiling            & Upturned mouth            & \textit{Zygomatic Major}                       & 12                &           \adjustimage{width=1.8cm,height=1cm,valign=m}{images/t7}          \\
Puckering          & Puckered lip corners      & \textit{Buccinator}                            & 14                &             \adjustimage{width=1.8cm,height=1cm,valign=m}{images/t8}         \\
Frowning           & Downturned mouth          & \textit{Triangularis}                          & 15                &            \adjustimage{width=1.8cm,height=1cm,valign=m}{images/t9}          \\ 
Lowering lower lip & Mouth shape for "s" sound & \textit{Depressor Labii Inferioris}            & 16                &            \adjustimage{width=1.8cm,height=1cm,valign=m}{images/t10}          \\ 
Shrugging          & Squashed lower lip        & \textit{Mentalis}                              & 17                &   \adjustimage{width=1.8cm,height=1cm,valign=m}{images/t11}                   \\
Pursing lips       & Mouth shape for kissing   & \textit{Orbicularis Oris}                      & 18                &    \adjustimage{width=1.8cm,height=1cm,valign=m}{images/t12}                  \\ 
Grimacing          & Stretched lower lip       & \textit{Risorius/Platysma}                     & 20                &      \adjustimage{width=1.8cm,height=1cm,valign=m}{images/t13}                \\
Lip loosening      & Mouth shape for "sh"      & \textit{Orbicularis Oris}                      & 22a-c             &    \adjustimage{width=1.8cm,height=1cm,valign=m}{images/t14}                  \\ 
Lip tightening     & Contracted mouth          & \textit{Orbicularis Oris}                      & 28a-c             &      \adjustimage{width=1.8cm,height=1cm,valign=m}{images/t15}                \\ 
Lip locking        & Lips pressed together     & \textit{Orbicularis Oris}                      & 24a-c             &       \adjustimage{width=1.8cm,height=1cm,valign=m}{images/t16}               \\ 
Moving jaw         & Open/Closed mouth         & \textit{Digastric, Temporalis, Masseter}       & 26a-d             &    \adjustimage{width=1.8cm,height=1cm,valign=m}{images/t17}                  \\ 
\bottomrule
\end{tabularx}

\end{table*}
\section{Overview of Active Snake Contours}
\label{sec:snakesOverview}
Active Snake Contours were developed by Kass et al.~\cite{snakes1988} in 1988. 
``Snakes'' are deformable models which lock onto features of an image 
through energy minimization. They are controlled by image forces 
and user-defined constraints (e.g. smoothness), and can be used to find
edges and suggestive contours in images. In our case, image
forces push the snakes towards darker regions and edges-- the weight of 
each is determined by tweakable values, as discussed in 
section~\ref{sec:imp_details}.

A snake is defined by a set of $n$ points $v_{0}$ to $v_{n-1}$. 
The Snake's energy can be represented by:
$$E_{snake}^* = \int_0^1 E_{snake}(\mathrm{v}(s)) \mathrm{d}s = \int_0^1 E_{internal}(\mathrm{v}(s)) + E_{external}(\mathrm{v}(s))\mathrm{d}s $$

\textit{Internal energy.} This represents properties inherent to 
the snake shape itself (smoothness and continuity).
$$E_{internal} = E_{smooth} + E_{cont}$$

Target values for snake shape properties are tweakable (see table~\ref{table:tweakables}).

\textit{External energy.} This represents the external forces that
the reference image poses on the snake. The formula for external energy is:
$$E_{external} = w_{brightness} E_{brightness} + w_{edge} E_{edge}$$
where:
\begin{itemize}
\item $E_{brightness}$ represents the energy resulting from the image's grayscale gradient.
\item $E_{edge}$ represents the energy resulting from edges in the image.
\item $w_{brightness}$ and $w_{edge}$ are relative weights.
\end{itemize}

Snake energy is iteratively minimized using gradient descent.

For our purposes, the inital guesses for the snakes are the contours of 
the input sketch, and the reference image is a render of a face model
generated as a middle step (see section~\ref{sec:overview}).

\section{Contributions}
This work makes the following contributions:
\begin{itemize}
\item Generating a dataset of 3D model/face sketch pairs using Valley Girl. This differs from
    the dataset by Cao et al.~\cite{cao2014} and Han et al.~\cite{deepSketch}, where the 3D models are extracted from real photos,
    in that all of our variations are random. Therefore, we are sure not to have any biases in gender or ethnicity.
\item FACS AUs are detected from face sketches for the first time in the literature. 
    We use our generated data to train classifiers to recognize 12 FACS AUs in face sketches. 
\item Generating a 3D model from the sketch using the recognized expression, Valley Girl, 
    and Active Snake Contours.
\end{itemize}

\section{Overview of method}
\label{sec:overview}

\begin{figure}[ht]
\centering
\includegraphics[width=6in]{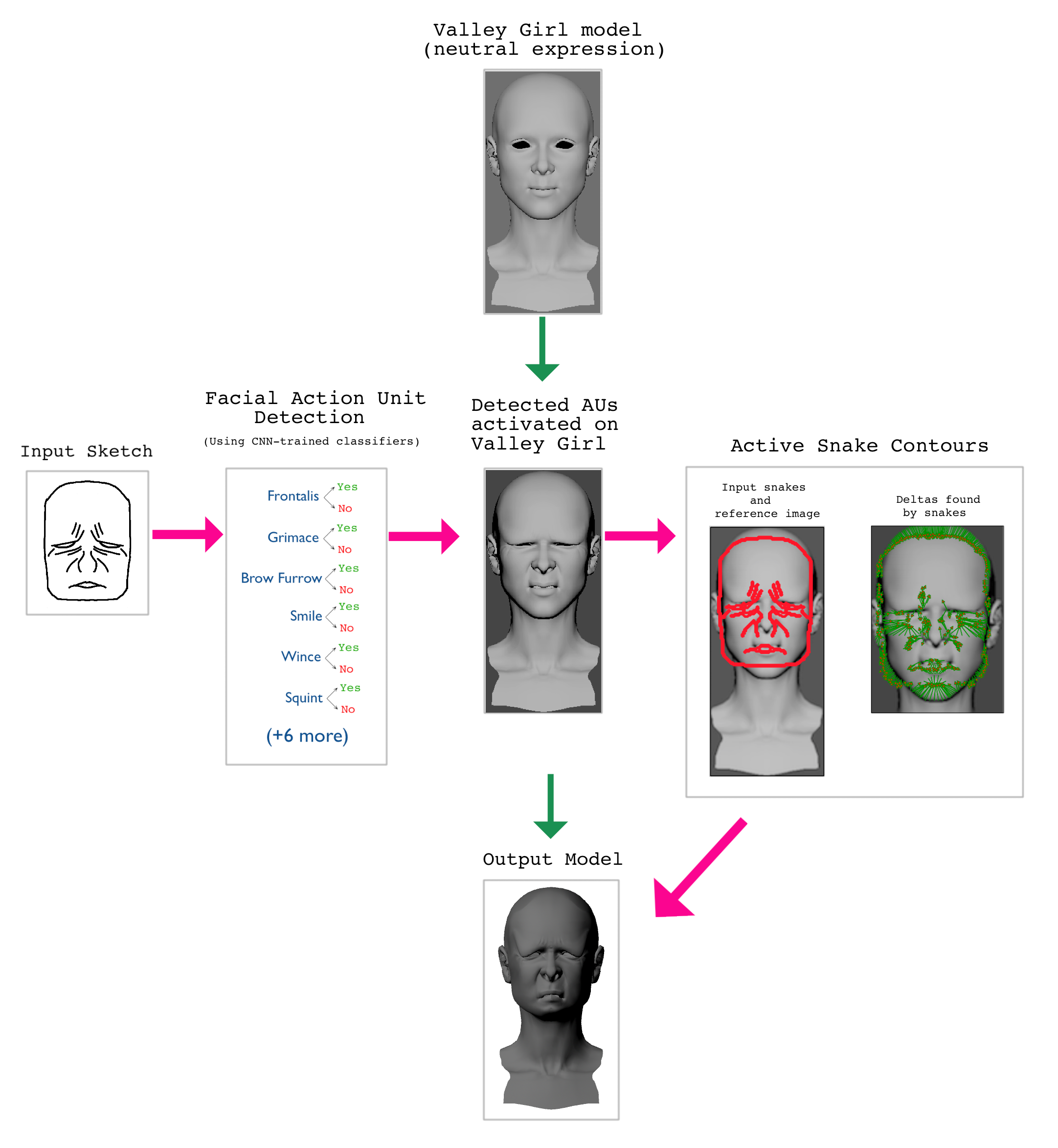}
\caption{High-level overview of method.}
\label{fig:method}
\end{figure}

\textit{Setup and training.} We have a parametric face 
model $F$ (Valley Girl) with scalar parameters representing race, gender, and FACs AUs activation levels,
and vectors representing head and neck rotation. 
Sketch/Model pairs are generated by posing $F$ through random assignment of values 
to its parameters. Once the face is posed, a corresponding sketch is generated through
non-photorealistic rendering. 
Convolutional Neural Networks are trained to partially recover the FACS AU parameters
from the sketch (chapter~\ref{ch:cnn}).

\textit{Input.} Sketch of a face.
\begin{enumerate}
\item The trained CNNs are used to recover the facial expression of 
the input face is detected by detecting the active facial actions.
\item The expression is then duplicated on the 3D model of Valley Girl~\cite{landreth}.
\item Gaps between the model from the previous step and the input sketch
are found using Active Snake Contours~\cite{snakes1988} and applied
on the model using soft transforms to produce the final result. 
For this step, user input is required to align the sketch with 
Valley Girl and set the head rotation if needed.
This step is covered in chapter~\ref{ch:snakes}.
This deformation can be applied to the parametric face model so 
that it can continue to be animated using its existing parameters.
\end{enumerate}

Figure~\ref{fig:method} illustrates the method.
Results are shown in chapter~\ref{ch:results}.

\section{Related Work}

\subsection{Generating 3D models from face sketches} 
\label{subsec:relatedSketches}
Most notably, Han et al.~\cite{deepSketch} developed DeepSketch2Face, an
interactive sketching system for generating 3D face models. They use an augmented
version of the dataset of 3D face models developed by Cao et al.~\cite{cao2014}. 
The dataset of 3D models is based on real photographs of varying expressions and 
identities. Sketches for each model are automatically generated, and the combined dataset of 
3D models and corresponding sketches is used to train a CNN 
basesd deep regression network to infer 3D models from 2D sketches. \\
This work differs from theirs in a number of ways: \\
\textit{Firslty}, despite the fact that the dataset by Cao et al.~\cite{cao2014} contains
both genders, all of the example outputs shown in ~\cite{deepSketch} look like males,
a shortcoming which they do not explain. It could be that the ratio of 
males to females in the dataset is too high. \\
\textit{Secondly}, Han et al.~\cite{deepSketch} and Cao et al.~\cite{cao2014} do not provide
sufficient details on the facial actions used to vary the expressions. Even though 
Cao et al.~\cite{cao2014} list some precise facial actions, e.g. ``lip puckerer'',
they also list ill-defined facial actions such as ``anger'' and ``sadness''. 
As discussed in section~\ref{sec:facs}, emotions can vary greatly in how they are expressed on the face. 
All of the facial actions used in this work,
based on FACS~\cite{ekman1977facial} and the work of Chris Landreth~\cite{landreth}, are precise. 
After the FACS AUs are detected, the expression can be accurately reproduced on Valley Girl (or any
parametrized face model that uses FACS). \\
\textit{Thirdly}, the outputs generated by Han et al.~\cite{deepSketch} veer more towards
caricatures than realistic-looking faces. This probably has to do with ``exaggeration'',
a method used by Cao et al.~\cite{cao2014} to vary the dataset by exaggerating all features in the face
simultaneously. This does not align with naturally-occurring faces; a person can have just one
exaggerated feature, or they can make an exaggerated expression without having exaggerated features.
Because we generate the data using Valley Girl, each parameter and facial 
action can be controlled separately, leading to more realistic looking models and sketches in the dataset. \\ 
\textit{Finally}, one of the limitations Han et al.~\cite{deepSketch} point out is that 
wrinkles cannot be created at novel locations (e.g. if the given sketch contains a scar 
at an unusual place, it will not be present in the output model). Even though we rely
on CNNs to recognize the expression in the sketch, our approach of using
Active Snake Contours to find a mapping between the contours in the sketch and the 3D face model
allows us to know when there's a contour that does not map to anything and that needs to be 
handled in another way, making it a step towards solving that limitation. 
Currently, these contours are excluded, but they are known.

Jiang et al.~\cite{sketch2010} deform an existing 3D model to match a sketch. 
First, they create ``sketchy face'', a version of the input sketch where 
the contours are processed to close gaps and remove unnecessary intersections-- 
they are then divided into Outer Contours and Inner Contours. After that, a mapping
between the contours of the model and the contours of the sketch is found using 
anatomical knowledge of the face (e.g. the nose of the model is divided into 
three parts and each is treated separately). Sucontphunt et al.~\cite{sketch2012}
explore a similar approach.\\
Our method also involves deforming an existing model to match a 
sketch, but we explicitly deal with facial expressions because they create 
many contours not found in a blank face. We also use a different approach to 
create the mapping between the model contours and the sketch contours, covered
in chapter~\ref{ch:snakes}.

\subsection{Generating 3D models from non-face sketches}

Inferring 3D Shape from a 2D sketch of an abstract shape is a popular 
topic of research in Computer Graphics. One of the earlier and most notable systems is
Teddy~\cite{igarashi1999teddy}. Another early approach is SmoothSketch, 
~\cite{2006smoothsketch}, where 
Karpenko et al.~\cite{2006smoothsketch} 
use Active Snake Contours in their approach, and so do
Kara et al.~\cite{kara2007sketch} who focus primarly on 
industrial sketches. Later approaches include True2Form~\cite{Xu:2014:True2Form},
which applies design and perception principles to reconstruct 3D curves 
from a design sketch.

Other solutions target sketches of specific types of objects and tend towards using
Procedural Modeling. For instance, Nishida et al.~\cite{nishida2016interactive}
focus on Urban modeling, and Longay et al.~\cite{longay2012treesketch} focus on 
Trees. 

More recently, Deep Learning approaches were applied to this problem.
Huang et al.~\cite{huang2016shape} use Convolutional Neural Networks
and procedural modeling to generate 3D models from sketches of 
abstract shapes. The CNN detects parameters that can be set on 
their procedural model, which is similar to our approach of detecting
FACS AUs (\ref{sec:facs}) and setting them on Valley Girl (\ref{sec:vg}). 
Their approach is meant to handle abstract shapes (i.e. it cannot handle faces) 
and can only produce outputs that the parameters of their model can be set
to produce. \\ 
Our approach explicitly handles faces and expands on just parameter 
detection and procedural modeling by
using Snakes (\ref{sec:snakesOverview}) to do contour matching between the 
model after parameters have been detected and set, and the input. This allows
us to apply additional transforms to the model and generate models that
go beyond what the parameters of Valley Girl can generate. 

\subsection{Detecting expressions from face sketches}
At the time of writing, I am only aware of one recent work in the literature that attempts
to recognize the facial expression in hand-drawn sketches. 
Moetesum et al.~\cite{moetesum2017sketch} train an SVM classifier to categorize
expressions in face sketches as ``happy'', ``sad'', ``angry'' and ``neutral''.
This is very limited and does not even cover the 6 basic human emotions~\cite{ekman1999basic}. 

Detecting expressions in photographs is more popular. 
Typically, the goal is to categorize the expression into one of the 6 basic emotions: 
``happiness'', ``sadness'', ``disgust'', ``anger'', ``surprise'' and ``fear''
~\cite{ekman1999basic}~\cite{Ma2004kv}~\cite{khandait2012automatic}~\cite{bu2005}. 
This may be suitable for some applications, but there is too much variation within each of 
the basic emotions to be able to accurately replicate an expression based just on its category,
as discussed in section~\ref{sec:facs}. In this work, we aim to recognize the constituent 
FACS Action Units ~\cite{ekman1977facial}.

Some attempts have been made to recognize AUs in photographs, 
and they mostly focus on facial feature extraction and analysis~\cite{tian2001recognizing}.

\subsection{Other Related Works}
\textit{Generating photographs from face sketches.}
Generating a realistic 3D model of the face instead can also
accomplish the generation of photographs when textures are added 
and the model is rendered. \\
Tang et al.~\cite{Wang:2009cp} synthesize photos from sketches 
in a patch-based manner using a dataset of photo-sketch pairs; 
for each patch in the sketch, a similar patch is located in the 
dataset, and its corresponding photo patch is used to estimate 
the patch in the photo being synthesized. \\
While this approach yields realistic-looking photos, 
only front-view is obtained, and the user
is not at liberty to choose the hairstyle, skin tone, and eye 
color of the output face (that said, photoshop can be used to 
make those changes). 

\textit{Using face sketches for suspect identification.}
Jain et al. ~\cite{Klare:2014io} tackle the problem of 
suspect identification using a set of facial attributes that attempt to capture 
all describable facial features, and using those attributes to search a database 
of faces without relying on a sketch artist. Another common approach is to use sketch-based 
face recognition~\cite{Pramanik:2012fz}, where the sketch is used to 
find possible matches from a database of faces. While those methods have their 
advantages, neither of them produces a photo or a 3D model, meaning that they
cannot be used to produce a search warrant based on the sketch, for example.
\chapter{Using Deep Learning to detect the expression in the given sketch}
\label{ch:cnn}

\section{Dataset}
\label{sec:dataset}
The dataset is comprised of over 21,000 pairs of 3D models and their 
corresponding sketches, randomly generated 
using Valley Girl and Suggestive Contours~\cite{decarlo2003suggestive}.\\
Even though using generated data might introduce some biases related to the 
model used and the technique for generating the sketch, 
deep learning models trained on synthetic data can outperform models trained 
on real data when tested on real data~\cite{7299105}. This might be due to the
guaranteed randomization in generated data. As discussed in 
subsection~\ref{subsec:relatedSketches}, we avoid biases in gender or ethnicity,
which datasets collected from real photographs are likely to suffer from.

In order to generate the sketches, the following parameters were randomized:

\begin{itemize}
\item \textit{Facial actions}. e.g. frowning, brow furrowing, smiling, squinting, etc.
For each facial action, an integer activation value is chosen at random between 0 and 5. 
When manipulating the model manually, it's reasonable to go higher than that 
(technically the values go up to 10, but start to look unrealistic around 8). 
However, if due to random generation several of the parameters are set too high, 
the model would end up looking deformed, which is why the maximum value was chosen
as such. Table~\ref{table:levels} shows the 6 levels of activation for ``frontalis'' Action Unit.
\item \textit{Head rotation}. Head rotation is randomized around all 3 axes (independently). 
For each axis, a random value is chosen from $[0,max]$. 
However, the model is constrained to lie in the $3/4$ to full frontal view.
\item \textit{Facial features}. The facial features were varied through direct randomization
of features such as eye size, eye slant, head shape, etc, 
and also through the randomization of gender and ethnicity. 
\item \textit{Level of detail in the generated sketch}. To generate a sketch from the generated model, 
contours and suggestive contours are extracted~\cite{decarlo2003suggestive}. Ridges, valleys 
and, suggestive ridges can be included or excluded to vary the level of detail; 
3 levels of detail were simulated in the generated sketches.
\end{itemize}

In total, 53 parameters were randomized to generate each 3D model/sketch pair. 

\begin{table*}[t]
\centering
\caption{Sketches corresponding to activation levels 0-5 for the ``frontalis'' Action Unit,
 shown on an otherwise neutral Valley Girl face. Activation level 0 is considered a negative sample,
 while the rest are considered positive samples.}
\label{table:levels}
\begin{tabularx}{\textwidth}{C|CCCCCC}
\toprule
\textbf{Activation Level} & 0 & 1 & 2 & 3 & 4 & 5 \\ 
\midrule
& \adjustimage{width=2cm,height=2.73cm,valign=m}{images/frontalis/0} 
& \adjustimage{width=2cm,height=2.73cm,valign=m}{images/frontalis/1} 
& \adjustimage{width=2cm,height=2.73cm,valign=m}{images/frontalis/2} 
& \adjustimage{width=2cm,height=2.73cm,valign=m}{images/frontalis/3} 
& \adjustimage{width=2cm,height=2.73cm,valign=m}{images/frontalis/4} 
& \adjustimage{width=2cm,height=2.73cm,valign=m}{images/frontalis/5} 
\\
\bottomrule
\end{tabularx}
\end{table*}

\section{Training}
\label{sec:training}

For each facial action, we aim to determine whether it is active or not. 
To that end, we train one CNN for each facial action, where the output layer contains two nodes, 
one for ``yes'' and one for ``no''. A pre-trained version of GoogLeNet~\cite{Szegedy_2015_CVPR} is fine-tuned, 
with three layers added on top (a convolutional layer, a max pooling layer and a softmax layer). 

The softmax layer applies a softmax nonlinearity to the output, and calculates the cross-entropy 
between the predictions and the label index-- i.e. the cross-entropy loss function measures 
how closely the model's predictions match the target labels (``yes''/``no'')~\cite{tensorflowTut}. 
Cross-entropy loss is most commonly preferred to other loss functions 
(e.g. squared error) for classification tasks, where each output node represents 
a different hypothesis about the input image. Like log loss, cross entropy heavily penalizes
predictions which are both confident and incorrect~\cite{Goodfellow-et-al-2016}. 
For classification problems involving only two labels, the cross entropy 
is $-(ylog(p) + (1-y)(log(1-p))$, where $y$ is the correct label index (0 or 1), and 
$p$ is the model's predicted probability.

For training each model, the mean cross entropy is minimized using 
10,000 iterations of stochastic gradient descent, with a learning rate of 0.01. 
Gradient descent is an optimization algorithm that minimizes the loss value (mean
cross entropy) by iteratively moving in the direction of steepest descent as defined
by the negative of the gradient.

The dataset was split as such: 80\% for training, 10\% for validation and 10\% for testing.

Classifiers were trained for 12 facial actions: frontalis, medialis frontalis, brow furrow,
grimace, grimace (left), grimace (right), sneer, wince, squint, outer squint, smile, and pucker. 
Tables~\ref{table:samples1} and~\ref{table:samples2} shows example ``yes'' samples for these
facial actions (excluding the asymmetrical grimaces). These facial actions were chosen to test
how well the detection would work on all facial actions. For instance, the chosen facial actions
include asymmetries (grimace (right) and grimace (left)) and subtle differences between actions
(e.g. squint and outersquint, or frontalis and medialis frontalis). This is meant to provide
an idea of how much the detection can be extended in future work (as discussed in~\ref{sec:future} 
I believe the detection can be expanded to all FACS AUs).

To train each one, the dataset is scanned and split into yes/no samples-- an Action Unit activation
level of 0 is considered a negative sample for that AU, while activation levels 1-5 are considered
positive samples. All samples were used in training each of the 12 classifiers.

Training was done using Tensor Flow~\cite{tensorflow2016}, with hardware
acceleration support using NVIDIA Quadro and CUDA~\cite{nvidia2010programming}.

\begin{table*}[t]
\centering
\caption{Examples of ``yes'' samples for the Facial Actions: Brow Furrow, 
Medialis Frontalis, Frontalis, Squint, Outer Squint, and Smile.}
\label{table:samples1}
\begin{tabularx}{\textwidth}{L{1}|L{1} L{1} L{1} L{1}}
\toprule
\textbf{Facial Action Unit} &  &  &  & \\ 
\midrule
Brow Furrow
& \adjustimage{width=2.5cm,height=3.82cm,valign=m}{images/samples/brow_furrow/1} 
& \adjustimage{width=2.5cm,height=3.82cm,valign=m}{images/samples/brow_furrow/2}  
& \adjustimage{width=2.5cm,height=3.82cm,valign=m}{images/samples/brow_furrow/3} 
& \adjustimage{width=2.5cm,height=3.82cm,valign=m}{images/samples/brow_furrow/4}  \\

Medialis Frontalis
& \adjustimage{width=2.5cm,height=3.82cm,valign=m}{images/samples/med_frontalis/1} 
& \adjustimage{width=2.5cm,height=3.82cm,valign=m}{images/samples/med_frontalis/2} 
& \adjustimage{width=2.5cm,height=3.82cm,valign=m}{images/samples/med_frontalis/3} 
& \adjustimage{width=2.5cm,height=3.82cm,valign=m}{images/samples/med_frontalis/4}  \\
  
Frontalis
& \adjustimage{width=2.5cm,height=3.82cm,valign=m}{images/samples/frontalis/1} 
& \adjustimage{width=2.5cm,height=3.82cm,valign=m}{images/samples/frontalis/2} 
& \adjustimage{width=2.5cm,height=3.82cm,valign=m}{images/samples/frontalis/3} 
& \adjustimage{width=2.5cm,height=3.82cm,valign=m}{images/samples/frontalis/4}  \\

Squint
& \adjustimage{width=2.5cm,height=3.82cm,valign=m}{images/samples/squint/1} 
& \adjustimage{width=2.5cm,height=3.82cm,valign=m}{images/samples/squint/2} 
& \adjustimage{width=2.5cm,height=3.82cm,valign=m}{images/samples/squint/3} 
& \adjustimage{width=2.5cm,height=3.82cm,valign=m}{images/samples/squint/4}  \\

Outer Squint
& \adjustimage{width=2.5cm,height=3.82cm,valign=m}{images/samples/outer_squint/1} 
& \adjustimage{width=2.5cm,height=3.82cm,valign=m}{images/samples/outer_squint/2}  
& \adjustimage{width=2.5cm,height=3.82cm,valign=m}{images/samples/outer_squint/3}  
& \adjustimage{width=2.5cm,height=3.82cm,valign=m}{images/samples/outer_squint/4}  \\

Smile
& \adjustimage{width=2.5cm,height=3.82cm,valign=m}{images/samples/smile/1} 
& \adjustimage{width=2.5cm,height=3.82cm,valign=m}{images/samples/smile/2} 
& \adjustimage{width=2.5cm,height=3.82cm,valign=m}{images/samples/smile/3}  
& \adjustimage{width=2.5cm,height=3.82cm,valign=m}{images/samples/smile/4}  \\

\bottomrule
\end{tabularx}
\end{table*}

\begin{table*}[t]
\centering
\caption{Examples of ``yes'' samples for the Facial Actions: Wince, Pucker, Sneer, and Grimace.}
\label{table:samples2}
\begin{tabularx}{\textwidth}{L{1}|L{1} L{1} L{1} L{1}}
\toprule
\textbf{Facial Action Unit} &  &  &  & \\ 
\midrule
Wince
& \adjustimage{width=2.5cm,height=3.82cm,valign=m}{images/samples/wince/1} 
& \adjustimage{width=2.5cm,height=3.82cm,valign=m}{images/samples/wince/2}  
& \adjustimage{width=2.5cm,height=3.82cm,valign=m}{images/samples/wince/3} 
& \adjustimage{width=2.5cm,height=3.82cm,valign=m}{images/samples/wince/4}  \\

Pucker
& \adjustimage{width=2.5cm,height=3.82cm,valign=m}{images/samples/pucker/1} 
& \adjustimage{width=2.5cm,height=3.82cm,valign=m}{images/samples/pucker/2} 
& \adjustimage{width=2.5cm,height=3.82cm,valign=m}{images/samples/pucker/3}  
& \adjustimage{width=2.5cm,height=3.82cm,valign=m}{images/samples/pucker/4}  \\
  
Sneer
& \adjustimage{width=2.5cm,height=3.82cm,valign=m}{images/samples/sneer/1} 
& \adjustimage{width=2.5cm,height=3.82cm,valign=m}{images/samples/sneer/2} 
& \adjustimage{width=2.5cm,height=3.82cm,valign=m}{images/samples/sneer/3} 
& \adjustimage{width=2.5cm,height=3.82cm,valign=m}{images/samples/sneer/4}  \\

Grimace
& \adjustimage{width=2.5cm,height=3.82cm,valign=m}{images/samples/grimace/1} 
& \adjustimage{width=2.5cm,height=3.82cm,valign=m}{images/samples/grimace/2} 
& \adjustimage{width=2.5cm,height=3.82cm,valign=m}{images/samples/grimace/3}  
& \adjustimage{width=2.5cm,height=3.82cm,valign=m}{images/samples/grimace/4}  \\

\bottomrule
\end{tabularx}
\end{table*}

\subsection{Clarification of subtle differences between some Facial Actions}

This is to make it easier for the reader to understand the examples shown in
tables~\ref{table:samples1} and~\ref{table:samples2}.

\textit{Wince and Sneer}. Wince and Sneer both create wrinkles downwards from the nostrils, 
which might be confusing. The difference is that for Wince, muscles of the nose are used, 
whereas for Sneer, muscles of the mouth are used. Sneer tends to shape the upper lip into
a swiggly line. 

\textit{Frontalis and Medialis Frontalis}. Frontalis creates horizontal wrinkles across 
the entire forehead, whereas Medialis Frontalis only creates wrinkles in the center
of the forehead and can shape the wrinkles and brows into a subtle ``V''.

\textit{Squint and Outer Squint}. As suggested by the name, Outer Squint mostly affects
the outer corner of the eyes. It tends to create crow's feet going upwards from the 
outer corner of the eye. Squint makes the eyes smaller in a much more pronounced way
and exaggerates the crow's feet wrinkles going downwards from the outer corner of the eye.
\section{Evaluation}
\label{sec:cnnEval}

\begin{figure}[ht]
\centering
\includegraphics[width=4in]{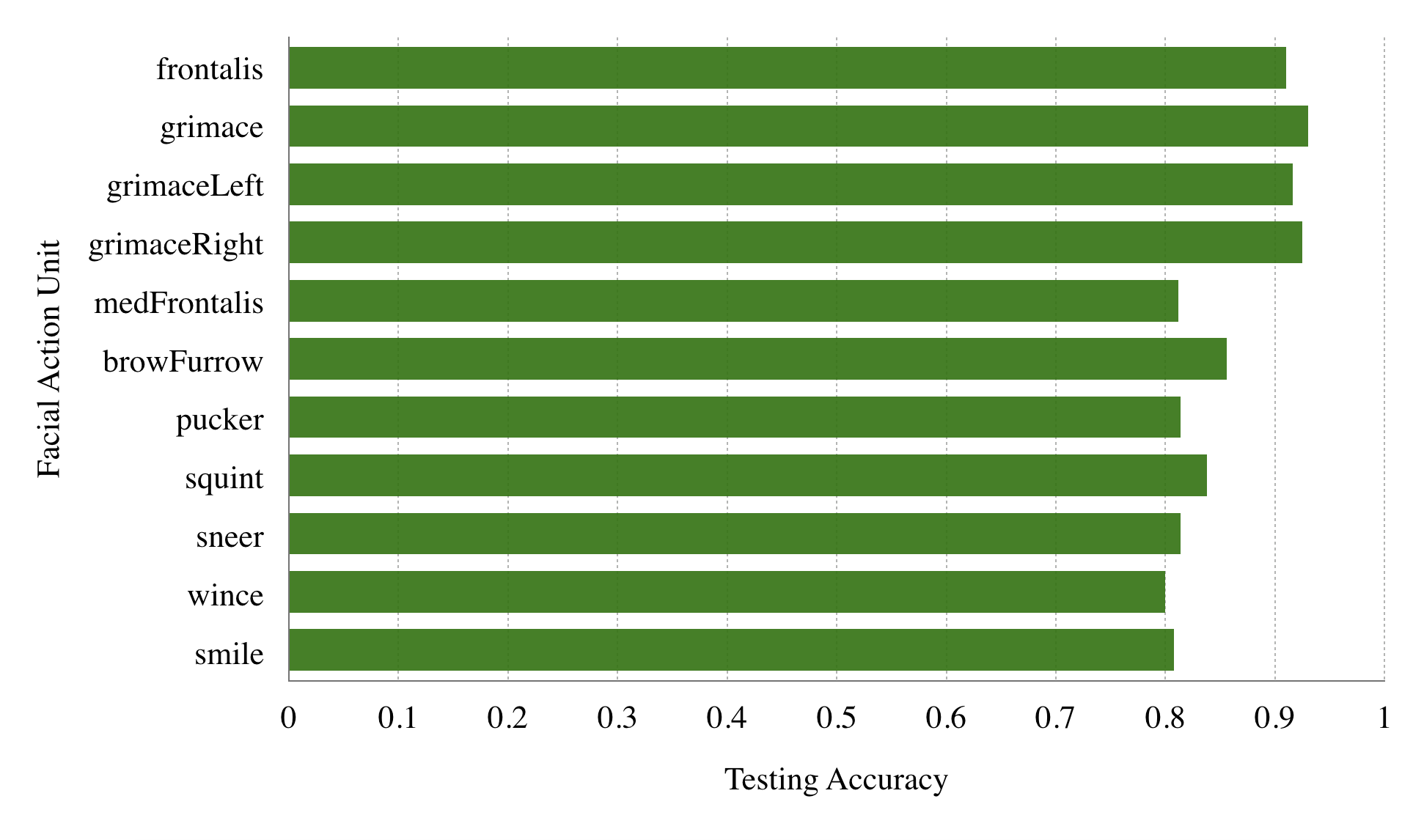}
\caption{Accuracy of the 11 trained AU classifiers.}
\label{fig:results}
\end{figure}

\begin{figure}[ht]
\centering
\includegraphics[width=3in]{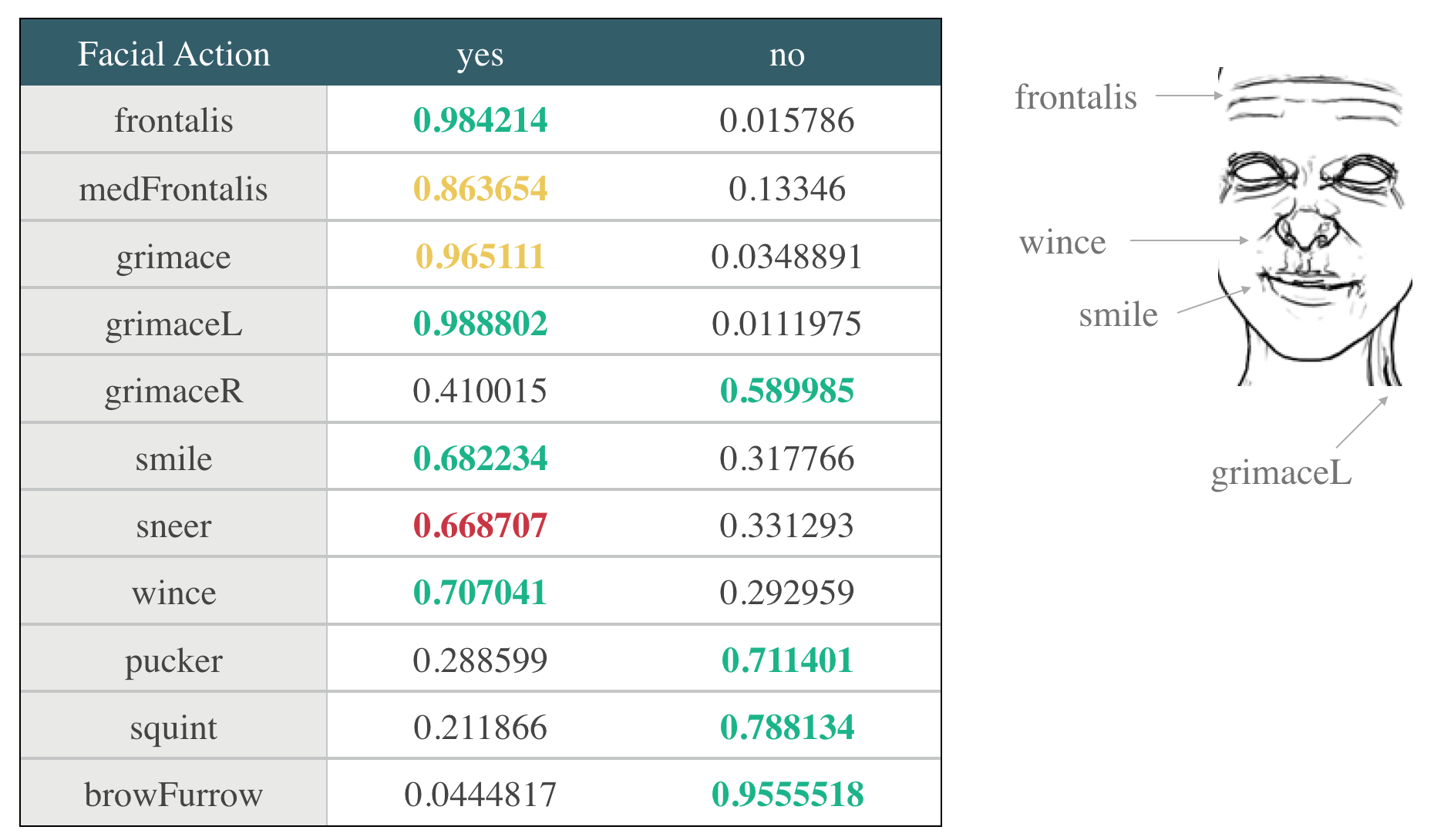}
\caption{AU detection example 1, showing values at the output nodes for each 
classifier. Correct results are in green. Arguably correct results
are in orange and are discussed in section~\ref{sec:cnnEval}.}
\label{fig:ex1}
\end{figure}

\begin{figure}[ht]
\centering
\includegraphics[width=3in]{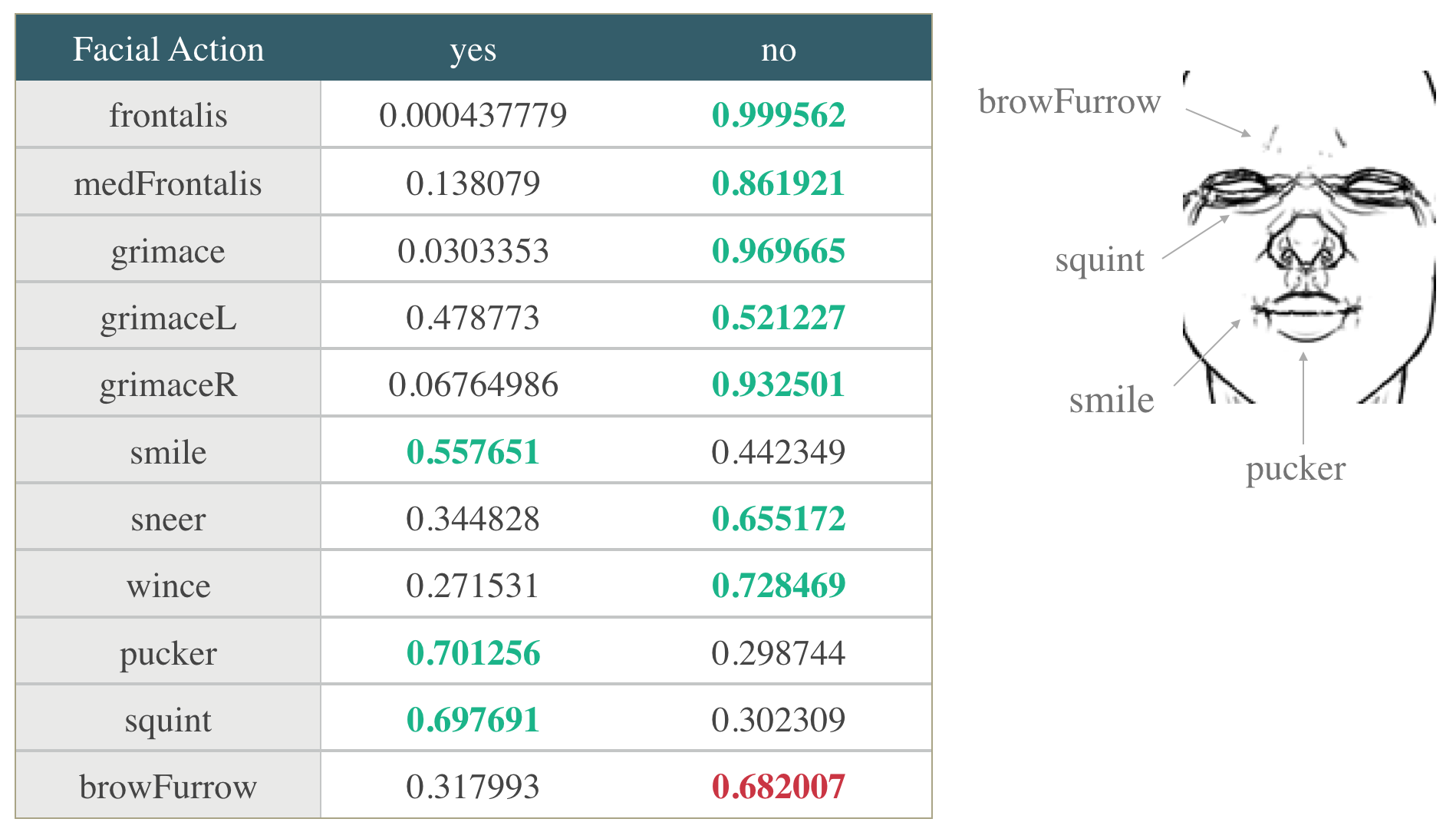}
\caption{AU detection example 2, showing values at the output nodes for each 
classifier. Correct results are in green. Arguably correct results
are in orange and are discussed in section~\ref{sec:cnnEval}.}
\label{fig:ex2}
\end{figure}

Figure~\ref{fig:results} shows the accuracy of each of the trained classifiers as tested on the 10\% of the samples 
set aside for testing. Figures~\ref{fig:ex1} and~\ref{fig:ex2} show two detailed examples using 
manually-constructed expressions. For figure~\ref{fig:ex1}, ``medFrontalis'' and ``grimace'' are detected as active, 
but these are arguably acceptable mistakes. The \emph{medialis frontalis} muscle is anatomically part of the \emph{frontalis}, 
which is active. 
The other dubious result is that ``grimace'' was detected when only ``grimaceL'' was preset. One possible solution would be
to remove samples where ``grimaceL'' and ``grimaceR'' are active from the ``yes'' samples for grimace. (To be clear,
they are currently included only when ``grimace'' (symmetrical) is also active on top of them.) Another possible 
solution is that in cases where right/left versions exist of the facial action, 
only the one with the highest score/probability would be used. 

Chapter~\ref{ch:results} provides examples using hand-drawn sketches.

\chapter{Using Active Snake Contours to close the gaps between the 3D model with the reproduced expression and the sketch}
\label{ch:snakes}

At this stage, we have the input sketch, and can easily generate a 3D face model 
with the same expression by setting the FACS AU parameters detected as described 
in chapter~\ref{ch:cnn} on Valley girl (see sections~\ref{sec:vg} and~\ref{sec:facs}). 
Replicating the expression is not enough for our model to match the sketch-- 
we now need to morph the model to match the features of the face in the sketch. This
setup lends itself very well to Active Snake Contours~\cite{snakes1988} 
(see section~\ref{sec:snakesOverview}).

\section{Method}
\label{sec:method}
\textit{Inputs.} The input sketch, and a list of AUs detected by 
the CNNs which were trained as described in chapter~\ref{ch:cnn}.
\begin{enumerate}
\item The detected AUs are activated on Valley Girl, and a grayscale shaded 
    image is produced of the result. (RTSC~\cite{decarlo2003suggestive} is 
    used for this, but it could also be a render.) I will refer to this
    as the Valley Girl expression render.
\item User input is needed to align the sketch with the VG expression render, 
    then both images are cropped and resized to 91x200 pixels.
\item The sketch is converted to binary, gaps are closed, the outline is
    thinned~\cite{nicolescu2014parallel} (optional, depends on how 
    aggressive the gap-closing was), and contours are extracted using the Marching Squares
    algorithm, which is a special case of Marching Cubes~\cite{lorensen1987marching}. 
    Contours that are too short (i.e. 1 point) are discarded.
\item Each of the extracted contours is run as a Snake and mapped onto
    the VG expression render. 
\item For each point on each output snake (coordinates on model), the delta 
    from the input snake is recorded. If many snake points landed on the
    same target, the deltas are averaged. This is to ensure that one vertex
    will not have the same change re-applied to it due to high contour density,
    for example. \\ 
    Deltas that are too large are discarded; these probably come
    from contours that were either: incorrectly mapped, or that do not exist
    in the model. 
\item A mapping is created between the VG vertex locations and pixel locations. 
    The conversion between vertex locations and pixel locations is done using
    the camera's projection and post-projection matrices, and the model's inclusive
    matrix. Front view is always used for the camera.  
    In addition, the y-coordinate needs to be flipped due to the different 
    coordinate systems in Maya and NumPy. 
    The coordinates are scaled according to the image size
    and the bounding box of the model.
\item Then, for each pixel which has a delta, the corresponding vertex 
    is found using the mapping. Vertices with higher Z values can be preferred
    using a tweakable value (see table~\ref{table:tweakables}).
    The delta is then scaled according to the image size and bounding box of the
    mesh, and the transform is applied using Soft Select.

    It's important to note that the mapping between vertices and pixels is done one
    time at the start. Doing this on the fly (i.e. after transforms from each iteration
    have been applied) leads to undesirable results.
\end{enumerate}

Refer to table~\ref{table:example_sphere} for a simple example demonstrating the idea
behind steps 3--7. Real examples are provided in chapter~\ref{ch:results}.

\begin{table*}[t]
\centering
\caption{Simple example showing how Active Snake Contours are used to 
    apply transformations from a sketch to the 3D Sphere model.}
\label{table:example_sphere}
\begin{tabular}{cccc}
\toprule
\textbf{Drawing} & \textbf{Model} & \textbf{Snake Contours} & \textbf{Pixel Deltas}\\ 
\midrule
    \adjustimage{width=2.5cm,height=2.85cm,valign=m}{images/example_sphere/drawing} 
& \adjustimage{width=2.5cm,height=2.85cm,valign=m}{images/example_sphere/input_model} 
& \adjustimage{width=2.5cm,height=2.85cm,valign=m}{images/example_sphere/snakes} 
& \adjustimage{width=2.5cm,height=2.85cm,valign=m}{images/example_sphere/deltas_zoom} \\
\hline

\toprule
\textbf{Output (ssd=0.6)} & \textbf{Output (ssd=0.8)} & \textbf{Output (ssd=1.0)} \\
\midrule
    \adjustimage{width=2.5cm,height=2.6cm,valign=m}{images/example_sphere/output_0-6ssd}
& \adjustimage{width=2.5cm,height=2.6cm,valign=m}{images/example_sphere/output_0-8ssd}
& \adjustimage{width=2.5cm,height=2.6cm,valign=m}{images/example_sphere/output_1ssd}
& \\
\bottomrule

\end{tabular}
\end{table*}
\section{Implementation Details}
\label{sec:imp_details}

\textit{Software.} Implemented as a Maya Python script using NumPy, SciPy,
Scikit-Image, and Matlab packages. 

\textit{Tweakables.} Table~\ref{table:tweakables} shows the list of values that the user can tweak, 
and the default values used for each. Default values were used for the examples
shown unless otherwise specified. As the table mentions, the default value used for
Snake Max Step is 1 pixel-- it was later observed that lower values (e.g. 0.3) can give
better results. \\
For the snake type, ``free'' snakes were chosen by default even though ``fixed'' snakes
lead to more neat-looking output snakes (i.e. due to external restrictions, you will
not see the snake go wackily out of place when the end-points are fixed). However, choosing
``fixed'' would mean that even if the exact same contour existed in the image slightly to the 
left or slightly to the right, only the middle part of the snake would be free to 
move towards it, which led to results that were too subtle and did not match the sketch. 
Hence, ``free'' snakes were preferred, and the ones that veered too far off were not taken
into consideration.

\begin{table*}[t]
\centering
\caption{Tweakable values for extracting the transforms from the input sketch using Active Snake Contours
 and applying them to Valley Girl with the reproduced expression.}
\label{table:tweakables}
\setlength\extrarowheight{2pt}
\begin{tabularx}{\textwidth}{ G{0.25} | G{0.1} | G{0.25} | G{0.4} }
\toprule
\textbf{Description} & \textbf{Abbr.} & \textbf{Default Value} & \textbf{Notes}\\ 
\midrule
Gap Closing Shape
& ---
& Square of side-length 2
& 
\\ \hline
Thinning Iterations
& ---
& 1
& 
\\ \hline
Snake Type 
& ---
& \textit{free}
& The type can be \textit{free}, \textit{fixed} (holds end-points in place), or \textit{periodic} (end-points must meet) 
\\ \hline
Snake Smoothness
& ---
& 1.0
& Higher values will lead to smoother results but will capture fewer details 
\\ \hline
Snake Time Stepping
& ---
& 2.0
& 
\\ \hline
Snake Max Step (in pixels)
& ---
& 1
& Lower numbers (e.g. 0.3) can lead to better results but are also more affected by any
  glitches in the reference image.
\\ \hline
Snake Max Iterations
& ---
& 3000
& In our tests, snakes reached convergence before the specified number of maximum iterations.
\\ \hline
Snake Convergence 
& ---
& 0.1
& Higher values are less demanding; for a value 
of 1.0 the snake hardly moves.
\\ \hline
Snake Brightness Weight 
& ---
& -5
& Negative values mean the snake is pushed towards darker regions.
Absolute value is the weight relative to Edge Weight.
\\ \hline
Snake Edge Weight 
& ---
& 1
& Position values mean the snake is pushed towards edges. 
Absolute value is the weight relative to Brightness Weight.
\\ \hline
Max Delta (in pixels)
& ---
& 15
& Depends on the size of the image, and should 
be inversely proportional to the level of detail in the image.
\\ \hline
Low Depth Preference
& ---
& 0
& Should be a value between 0 and 1.
\\ \hline
Soft Select Distance
& ssd 
& 1.0 
& Depends on the size of the model (needed to be lower 
for the sphere in figure~\ref{table:example_sphere}, for instance).
\\ \hline
Soft Select Curve
& ssc
& Linear
&
\\ \hline
Mirror Output 
& ---
& False
& Whether the output model should be mirrored to remove asymmetries.
\\
\bottomrule
\end{tabularx}
\end{table*}

\chapter{Results}
\label{ch:results}

This chapter presents four example walkthroughs and discusses the results.
The first example, shown in table~\ref{example1}, was generated in the same way
as the sketches in section~\ref{sec:dataset} (but was not part of the dataset
unless it was randomly generated). The other three examples, shown in figures
~\ref{example2},~\ref{example3}, and~\ref{example4} were hand-drawn in GIMP. 

\section{Detected AUs.} 
\textit{Example 1.} (Shown in table~\ref{example1}.) 
The AUs detected were accurate. 
However, having both ``grimace'' and ``grimaceL'' on top of each other
led to the grimace being over-exaggerated in the generated Valley Girl.
One solution to this problem: When replicating the expression on Valley
Girl, the intensity of AUs with all 3 versions (symmetrical, left, and right) 
would be scaled down if the others were also detected.

\textit{Example 2.} (Shown in table~\ref{example2}.) Outersquint, squint, pucker,
and sneer were detected. It's unclear why sneer was detected-- it could 
due to the exaggerated dent in the upper lip which causes the upper contour
to resemble that of a sneer. The other AUs detected are accurate in my opinion,
although ``smile'' was missing.

\textit{Example 3.} (Shown in table~\ref{example3}.) Pucker and wince were detected.
Even though ``wince'' might seem counterintuitive, it's easy to see why it was
detected-- the wrinkles beneath the eyes are more pronounced when wince is active, 
and I believe activating it on Valley Girl enhanced the output model
in terms of how closely it matches the sketch.

\textit{Example 4.} (Shown in table~\ref{example4}.) Brow Furrow, wince,
sneer, squint, and outersquint were detected, all of which I find to be
accurate with the exception of sneer. 

\section{Output Snakes and the associated transforms} 
Even though the output from running the snakes looks kind of messy 
(most pronounced in example 1, as seen in table~\ref{example1}), 
the next step (showing the pixel deltas) is a big improvement over that. 
This is due to the fact that the deltas are averaged for each target pixel 
(or each target vertex), and deltas that are too large are discarded (refer
to section~\ref{sec:method}).

The neck contours in example 3~(shown in table~\ref{example3}) were mapped
to inner neck lines as opposed to mapping to the outline of the neck itself,
a limitation that is further discussed in section~\ref{sec:limitations}.

In example 4, Snake Pixel Step was reduced to 0.1, edges and brightness were
given equal weights (1 and -1 respectively), and Low Depth Preference was
increased to 0.2. 

\section{Final Result} 
For all examples, I found the final result
to be realistic and to sufficiently match the sketch. For example 
3 (shown in table~\ref{example3}), I mirrored the sketch in GIMP, making it
pretty symmetrical, and it turned out to have a particularly good looking output, 
suggesting that perhaps asymmetries need to be dealt with more explicitly.
One idea is to tone down the soft-selection fall-off distance when a transform 
is not mirrored, in order to make the asymmetrical transform more localized
such that it doesn't end up overly distorting the face as a whole.

For example 4, mirroring was used to produce both the sketch and the model.
For sketches that don't have asymmetries, there are no downsides 
to mirroring.

\section{Performance} 
All tests were done using a relatively old 
Macbook Pro~\cite{macbook} \footnote{All throughout testing and debugging, 
I experienced a plethora of crashes in
Maya, suggesting that perhaps my laptop is not up to the task.}.
The average time taken to detect all the AUs is 23.33 seconds (about 1.9 seconds
per AU).
The average time taken to run the snakes and generate the pixel deltas is 4.10 
seconds. The average time taken to apply the transforms and generate the final
result is 8.13 seconds. Overall, the average is 35.56 seconds. 

\section{Presentation} 
Hair, eyeballs, and skin textures were added 
manually for the renders.

\begin{table*}[t]
\centering
\caption{Walkthrough from input sketch to final result (example 1).}
\label{example1}
\begin{tabular}{ccc}
\toprule
\textbf{Drawing} & \textbf{Detected AUs} & \textbf{Valley Girl with AUs} \\ 
\midrule
    \adjustimage{width=2.5cm,height=5.5cm,valign=m}{images/example1/drawing} 
& \begin{tabular}{@{}c@{}}
    smile (0.57), grimace (0.92) 
    \\ outersquint (0.79), squint (0.99)
    \\ wince (0.66), frontalis (0.90)
    \\ sneer (0.94), grimaceL (0.90)
    \\ medfrontalis (0.67)
    \end{tabular} 
& \adjustimage{width=2.5cm,height=5.5cm,valign=m}{images/example1/vg} \\
\hline

\toprule
\textbf{Snake Contours} & \textbf{Pixel Deltas} & \textbf{Output Model} \\
\midrule
    \adjustimage{width=2.5cm,height=5.45cm,valign=m}{images/example1/snake}
& \adjustimage{width=3cm,height=4.41cm,valign=m}{images/example1/deltas_zoom}
& \adjustimage{width=2.5cm,height=5.5cm,valign=m}{images/example1/model} \\
\hline

\toprule
\textbf{Render (front)} & \textbf{Render (side1)} & \textbf{Render (side2)} \\
\midrule
    \adjustimage{width=3.5cm,height=6.38cm,valign=m}{images/example1/render_front}
& \adjustimage{width=3.5cm,height=6.38cm,valign=m}{images/example1/render_side2}
& \adjustimage{width=3.5cm,height=6.38cm,valign=m}{images/example1/render_side} \\
\bottomrule
\end{tabular}
\end{table*}

\begin{table*}[t]
\centering
\caption{Walkthrough from input sketch to final result (example 2).}
\label{example2}
\begin{tabular}{ccc}
\toprule
\textbf{Drawing} & \textbf{Detected AUs} & \textbf{Valley Girl with AUs} \\ 
\midrule
  \adjustimage{width=2.5cm,height=5.5cm,valign=m}{images/example2/drawing} 
& \begin{tabular}{@{}c@{}}
    outersquint (0.75), squint (0.99)
    \\ pucker (0.80), sneer (0.74)
  \end{tabular} 
& \adjustimage{width=2.5cm,height=5.5cm,valign=m}{images/example2/vg} \\
\bottomrule

\toprule
\textbf{Snake Contours} & \textbf{Pixel Deltas} & \textbf{Output Model} \\
\midrule
  \adjustimage{width=2.5cm,height=5.45cm,valign=m}{images/example2/snake}
& \adjustimage{width=3cm,height=4.41cm,valign=m}{images/example2/deltas_zoom}
& \adjustimage{width=2.5cm,height=5.25cm,valign=m}{images/example2/model} \\
\bottomrule

\toprule
\textbf{Render (front)} & \textbf{Render (side1)} & \textbf{Render (side2)} \\
\midrule
  \adjustimage{width=3.5cm,height=4.81cm,valign=m}{images/example2/render_front}
& \adjustimage{width=3.5cm,height=4.81cm,valign=m}{images/example2/render_side2}
& \adjustimage{width=3.5cm,height=4.81cm,valign=m}{images/example2/render_side} \\
\bottomrule
\end{tabular}
\end{table*}

\begin{table*}[t]
\centering
\caption{Walkthrough from input sketch to final result (example 3).}
\label{example3}
\begin{tabular}{ccc}
\toprule
\textbf{Drawing} & \textbf{Detected AUs} & \textbf{Valley Girl with AUs} \\ 
\midrule
    \adjustimage{width=2.5cm,height=5.5cm,valign=m}{images/example3/drawing} 
& \begin{tabular}{@{}c@{}}
    pucker (0.62), wince (0.71)
    \end{tabular} 
& \adjustimage{width=2.5cm,height=5.5cm,valign=m}{images/example3/vg} \\
\bottomrule

\toprule
\textbf{Snake Contours} & \textbf{Pixel Deltas} & \textbf{Output Model} \\
\midrule
    \adjustimage{width=2.5cm,height=5.45cm,valign=m}{images/example3/snake}
& \adjustimage{width=3cm,height=4.41cm,valign=m}{images/example3/deltas_zoom}
& \adjustimage{width=2.5cm,height=5.5cm,valign=m}{images/example3/model} \\
\bottomrule

\toprule
\textbf{Render (front)} & \textbf{Render (side1)} & \textbf{Render (side2)} \\
\midrule
    \adjustimage{width=3.5cm,height=4.45cm,valign=m}{images/example3/render_front}
& \adjustimage{width=2.89cm,height=4.45cm,valign=m}{images/example3/render_side2}
& \adjustimage{width=3.23cm,height=4.45cm,valign=m}{images/example3/render_side} \\
\bottomrule
\end{tabular}
\end{table*}

\begin{table*}[t]
\centering
\caption{Walkthrough from input sketch to final result (example 4).}
\label{example4}
\begin{tabular}{ccc}
\toprule
\textbf{Drawing} & \textbf{Detected AUs} & \textbf{Valley Girl with AUs} \\ 
\midrule
    \adjustimage{width=2.5cm,height=5.5cm,valign=m}{images/example4/drawing} 
& \begin{tabular}{@{}c@{}}
    browfurrow (0.84), sneer (0.86)
    \\ wince (0.76), squint (0.98)
    \\ outersquint (0.89)
    \end{tabular} 
& \adjustimage{width=2.5cm,height=5.5cm,valign=m}{images/example4/vg} \\
\bottomrule

\toprule
\textbf{Snake Contours} & \textbf{Pixel Deltas} & \textbf{Output Model} \\
\midrule
    \adjustimage{width=2.5cm,height=5.45cm,valign=m}{images/example4/snake}
& \adjustimage{width=3cm,height=4.41cm,valign=m}{images/example4/deltas_zoom}
& \adjustimage{width=2.5cm,height=5.17cm,valign=m}{images/example4/model} \\
\bottomrule

\toprule
\textbf{Render (front)} & \textbf{Render (side1)} & \textbf{Render (side2)} \\
\midrule
    \adjustimage{width=3.5cm,height=6.82cm,valign=m}{images/example4/render_front}
& \adjustimage{width=3.5cm,height=5.97cm,valign=m}{images/example4/render_side2}
& \adjustimage{width=3.5cm,height=6.1cm,valign=m}{images/example4/render_side} \\
\bottomrule
\end{tabular}
\end{table*}

\chapter{Discussion}

\section{Limitations}
\label{sec:limitations}

\textit{Snakes detect local minima.}
Before running the Snakes (see chapter~\ref{ch:snakes}), the two faces need to be
aligned, which we use user input to accomplish. The features only need to be
roughly aligned. However, if the proportions of the face in the input sketch vary
greatly from the proportions of Valley Girl's face (even after applying the correct
expression), the mapping obtained from the Snakes will contain errors. For instance, 
if the input sketch's mouth is entirely below Valley Girl's mouth, all contours from 
the input sketch will be mapped to just the lower lip. This is because Active Snake 
Contours find the local minima. \\
That said, Valley Girl has the proportions of a very average face, meaning that the
problem is not likely to happen for typical realistic faces, especially since 
we rely on user input for the alignment.

\textit{Creating new features.} 
Because we obtain the output model by deforming 
an existing model, creating new features is problematic. For instance, if 
a sketch has three eyes instead of two, the third eye will not be created.
More realistically, sketches might contain scars or an unusual number of wrinkles,
which will not be created. Han et al.~\cite{deepSketch} raised the same concern.
Subsection~\ref{subsec:create_contours} discusses possible solutions.

\textit{Oversights in the dataset.} Even though at least 53 parameters were randomized,
there were some oversights which resulted in certain biases. For instance, the 
neck size was not changed, and there there were only three options for ethnicity, 
which is not sufficient. In addition, age was not accounted for. 
The good news is that these oversights are relatively 
easy to correct by augmenting Valley Girl.

\section{Future Work}
\label{sec:future}

\subsection{Improvements to the Dataset and AU Detection}
We have the capacity to greatly enhance and augment the dataset (see section~\ref{sec:dataset}).
Given the parameters we can randomize, we can generate at least $8 \times e^{38}$ different faces. 
For each of those, multiple sketches in different styles can be generated.

As pointed out in section~\ref{sec:limitations}, the number of options for ethnicity
should be increased, and parameters need to be added
for age and neck shape.

We also have the capacity to train more classifiers to recognize even more AUs. 
The dataset contains the activation values of all AUs.

As described in section~\ref{sec:dataset}, the maximum activation level for Valley Girl
(\ref{sec:vg}) parameters was chosen to be lower than the maximum value supported
by the model. 
This was to avoid model defomration if too many values randomly happened to be too high. 
One better alternative to this approach could be to instead semi-randomly 
generate a sparse vector of AU activation values (i.e. first randomly choose $n$ AUs to active,
and then randomly choose an unconstrained activation value for each). This would lead to more
drastic variety in the dataset and could improve the robustness of the trained classifiers.

These improvements are relatively simple extensions to the work already done.

\subsection{Performance}
Even though the performance is acceptable for a user application, 
waiting half a minute in an interactive application is 
not ideal. The bottleneck on performance is the AU detection, which
takes an average of 1.94 seconds per AU (for a total of 23.3 seconds). 
This could greatly benefit from hardware acceleration. 

TensorFlow~\cite{tensorflow2016} does support hardware acceleration, 
but the computer used for development did not~\cite{macbook} (hardware acceleration
was used during the training period only). That said,
it would be too demanding to require that anyone using our software 
has to setup TensorFlow with hardware acceleration enabled (which is not 
a trivial task). One alternative is to use a dedicated server with 
the required setup to run the classifers, and have the client 
application query it.  

\subsection{Creating a UI}
Currently, the software exists as three scripts (to detect AUs, run snakes, 
and apply transforms) where the first takes the 
face sketch as input and the others take the output of the previous 
stage as input-- the third produces the output model. 
Given that all scripts are written in Python or Maya
Python, the software can be packaged as an interactive Maya plugin to improve
ease-of-use and user-friendliness. This could stand to greatly
enhance the output model given that users will be able to iterate on 
their sketches more quickly.

After a UI is developed user studies can be done to evaluate the output
and the experience.

\subsection{Extending the Method to Create New Features}
\label{subsec:create_contours}

As mentioned in section~\ref{sec:limitations}, creating new features
(e.g. a third eye, wrinkles at unusual locations, etc) is not currently supported.
However, we are able to detect which contours do not contribute to 
the output (refer to section~\ref{sec:method}). One possible approach
is to query the user on what they want done with these contours-- 
they could choose to have the relevant region hollowed, embossed, etc. 

\subsection{Changing the expression of the generated model}
Currently, the model is generated as an OBJ file containing just the face model.
We could instead apply the transforms directly onto Valley Girl 
(as opposed to applying them to a static OBJ version of her), which 
would allow the user to use the Valley Girl sliders to change her expression. 

In fact, this could be used to quickly and significantly augment the dataset 
(see section~\ref{sec:dataset}) by adding the generated faces as identities 
that could be randomized. 

However, since Valley Girl's parameters are implemented using Blend Shapes 
it remains to be seen whether this will give satisfactory results (that look 
more like the sketch than Valley Girl after blending with her blendshapes).

\section{Conclusion}

This work develops and implements a novel approach to the problem of generating
3D models from simple face sketches which combines both deep learning and contour matching 
between an existing 3D model and the sketch. Motivated by the fact that facial expressions
significaly alter and shape the contours of the face, we first detect the facial 
expression in the sketch and replicate it on a parametric 3D model. We then use
Active Snake Contours to find the mapping between the contours of the modified
model and those of the sketch.

The results are realistic and good-looking; in a matter of minutes, the user can add hair,
textures, and eyeballs to render their sketch as a photograph or export the model for use
in other applications.

We have also contributed a generated dataset of 3D models and sketch pairs which contains
randomized expressions, head rotations, ethnicities, and gender.\\ 
This dataset was used in training Convolutional Neural Networks to detect 
facial actions--- this contribution allows us to replicate the expression 
from a face sketch onto any face model which
uses FACS. Works on detecting FACS AUs from face sketches did not previously exist
in the literature at the time of writing.

\addcontentsline{toc}{chapter}{Bibliography}
\bibliographystyle{plainurl}
\bibliography{thesis}


\end{document}